\documentclass{article}
\usepackage{spconf,amsmath,graphicx}
\usepackage{hyperref}

\usepackage{enumitem}
\setlist{nosep, leftmargin=14pt}

\usepackage{mwe} 

\newcommand{\ie}{\textit{i}.\textit{e}.}
\newcommand{\eg}{\textit{e}.\textit{g}.}

\usepackage{lipsum}

\let\OLDthebibliography\thebibliography
\renewcommand\thebibliography[1]{
	\OLDthebibliography{#1}
	\setlength{\parskip}{0.5pt}
	\setlength{\itemsep}{1pt plus 0.1ex}
}

\title{A Robust Framework of Chromosome Straightening with ViT-Patch GAN}
\name{\begin{tabular}{c}Sifan Song$^{1,3 *}$\thanks{* Contributed Equally}, Jinfeng Wang$^{1,3 *}$, Fengrui Cheng$^{1 *}$, Qirui Cao$^{1}$, Yihan Zuo$^{1}$, Yongteng Lei$^{1}$,\\
Ruomai Yang$^{1}$, Chunxiao Yang$^{2}$, Frans Coenen$^{3}$, Jia Meng$^{1}$, Kang Dang$^{4 \dagger}$, Jionglong Su$^{1 \dagger}$\thanks{$\dagger$ Corresponding Authors. kdang@voxelcloud.net.cn; Jionglong.Su@ xjtlu.edu.cn} \end{tabular}}

\address{\small $^{1}$ Xi'an Jiaotong-Liverpool University, Suzhou, China\\ \small $^{2}$ Suzhou Sano Precision Medicine Ltd., Suzhou, China \\ \small $^{3}$ University of Liverpool, Liverpool, UK \\ \small $^{4}$ VoxelCloud, Inc., Los Angeles, USA \\[-3.0ex]}
\begin{document}
%
\maketitle
\begin{abstract}
Chromosomes carry the genetic information of humans. They exhibit non-rigid and non-articulated nature with varying degrees of curvature. Chromosome straightening is an important step for subsequent karyotype construction, pathological diagnosis and cytogenetic map development. However, robust chromosome straightening remains challenging, due to the unavailability of training images, distorted chromosome details and shapes after straightening, as well as poor generalization capability. In this paper, we propose a novel architecture, ViT-Patch GAN, consisting of a self-learned motion transformation generator and a Vision Transformer-based patch (ViT-Patch) discriminator. The generator learns the motion representation of chromosomes for straightening. With the help of the ViT-Patch discriminator, the straightened chromosomes retain more shape and banding pattern details. The experimental results show that the proposed method achieves better performance on Fréchet Inception Distance (FID), Learned Perceptual Image Patch Similarity (LPIPS) and downstream chromosome classification accuracy, and shows excellent generalization capability on a large dataset.
\end{abstract}
\begin{keywords}
Chromosome Straightening,  Vision Transformer, Generative Adversarial Networks.
\end{keywords}

\section{Introduction}
\label{sec:intro}

In a normal human cell, there are 22 pairs of autosomes (Type 1-22) and one pair of heterosomes (Type X \& Y in male and two copies of Type X in female). By karyotype analysis, chromosome aberrations can be detected in the diagnosis of many genetic diseases, such as the Klinefelter syndrome~\cite{bojesen2007klinefelter} and specific cancers~\cite{albertson2003chromosome}. The banding patterns of chromosomes (unique light and dark stained bands) provide important evidence for the development of cytogenetic maps. Due to their non-rigid nature, condensed chromosomes exhibit varying degrees of curvature under the microscope. Therefore, chromosome straightening is an important upstream task for chromosome classification~\cite{zhang2018chromosome} and the subsequent karyotype construction and cytogenetic map development~\cite{artemov2018development}. 

Chromosome straightening has been studied for a long time and its development may be described in three stages. (i) Printed images of bent chromosome are physically cut and rearranged for straightening~\cite{stegniui1991systemic,stegnii1978chromosome}. (ii) Geometric algorithms are extensively designed based on chromosome micrographs for automatic straightening, which consist of two main categories: extraction of the medial axis~\cite{arora2017algorithm,jahani2012centromere,somasundaram2014straightening,xie2019statistical} and finding bending points~\cite{roshtkhari2008novel,sharma2017crowdsourcing,zhang2021chromosome}. However, these geometric methods may extract inaccurate medial axes when chromosomes have large widths. The methods of bending point localization often use a cut line to separate chromosome arms, leading to discontinuous banding patterns. (iii) A Single Chromosome Straightening Framework (SCSF)~\cite{song2021novel} using conditional generative adversarial networks (cGAN) is proposed to straighten chromosomes with uninterrupted chromosome banding patterns. However, it requires a large number of input image pairs for training a model for each chromosome. ChrSNet~\cite{zheng2022chrsnet} is proposed as a two-module framework with self-attention mechanism, but some edge details are not well-preserved. In addition, there are many studies on image deformation, such as medical image registration~\cite{alam2019challenges,arora2017comparative}. The image registration technique requires two real-world chromosomes with the identical shape details but different curvatures, which is almost impossible to obtain. Thus it also fails in this chromosome straightening task. Recently, First Order Motion Model (FOMM)~\cite{siarohin2019first} and PCA-based Motion Estimation Model (PMEM)~\cite{siarohin2021motion} are proposed by learning key point-based and region-based representations for motion transfer, respectively. However, they still require to train on sufficient image pairs for robust performance. 

\begin{figure*}
	\centering
	\includegraphics[width=0.8\linewidth]{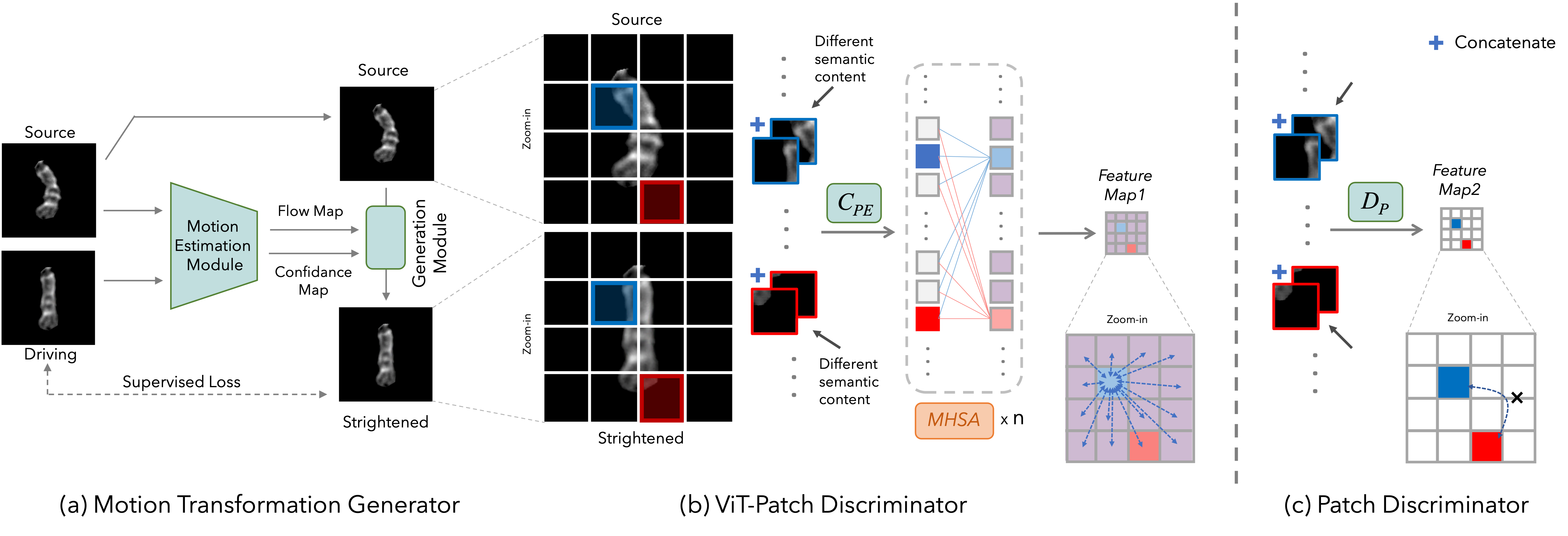}
	\vspace{-3pt}
	\caption{\footnotesize Overview of the architectures of the proposed framework, including (a) motion transformation generator and (b) our proposed ViT-Patch discriminator; (c) gives the structure comparison of a basic patch discriminator.} \label{Fig_vitgan}
\vspace{-5pt}
\end{figure*}

Challenges remain in the three stages of chromosome straightening. (1) \textit{Lack of of training images.} It is almost impossible to take micrographs of two chromosomes with identical stained banding pattern but different curvatures due to the diversity of random mutations, chromosome condensation and laboratory conditions. Thus, it is challenging to train a robust deep learning-based model for straightening. (2) \textit{Distorted chromosome details and shapes after straightening.} The condensed chromosomes are non-rigid with varying degrees of curvature. Straightened chromosome results require a high degree of preservation of consistent shape and details in the source image. Image registration and motion transfer methods tend to generate distorted results with driving image shapes. (3) \textit{Poor generalization capability.} The recent cGAN-based chromosome straightening framework~\cite{song2021novel} only learns mapping dependencies for each specific banding patterns, making it very time-consuming for large-scale applications.

The main contributions of this research address the above problems and are as follows: 
\begin{itemize}
	\item We propose a robust cGAN-based framework for chromosome straightening on a small dataset, by transfering the chromosome straightening task to the motion transformation task of non-rigid objects. 
	\item We propose a novel architecture, ViT-Patch discriminator, to improve the detail preservation ability of our framework. Compared with existing methods, straightened chromosomes retain more shape and banding pattern details of the corresponding source images. 
	\item Different from SCSF~\cite{song2021novel}, our trained model shows excellent generalization capability and has the ability to be applied to a large chromosome dataset for straightening. 
\end{itemize}

\vspace{-7pt}
\section{Methodology}
\vspace{-4pt}

\subsection{Network Architecture}
\vspace{-2pt}
Fig. \ref{Fig_vitgan} presents an overview of the proposed ViT-Patch GAN architecture. It is comprised of two parts, a motion transformation generator (Fig. \ref{Fig_vitgan}-a) and a ViT-Patch discriminator (Fig. \ref{Fig_vitgan}-b). Compared to SCSF~\cite{song2021novel}, ViT-Patch GAN requires a small dataset size for training (642 images). It straightens chromosomes with highly preserved details, and has high generalization capacity.

\textbf{Generalized Framework of Chromosome Straightening.} We apply PMEM~\cite{siarohin2021motion} as the generator part since it learns motion representations through self-learned region estimation. We can consider the chromosome straightening task as a motion transformation task for non-rigid objects. In the training stage, the generator requires a \textit{training source} and \textit{training driving} images with the same chromosome but different curvatures. As shown in Fig. \ref{Fig_vitgan}-a, using the Motion Estimation Module, the flow and confidence maps containing a combination of region and background transformations are then fed into the Generation Module to synthesize the straightened chromosome. Subsequently, the straightened chromosome is supervised with the \textit{training driving} image by a supervised reconstruction loss ($\mathcal{L}_1$), consistent with PMEM~\cite{siarohin2021motion}. One of the challenges of this task is the lack of \textit{training driving} data. However, PMEM was trained on sufficient image pairs (video clips containing only the same object)~\cite{siarohin2021motion}. On a small dataset, PMEM may inadequately transfer the shape of the driving image to the source image, leading to inaccurate straightening results. 

\begin{figure*}
	\centering
	\includegraphics[width=0.8\linewidth]{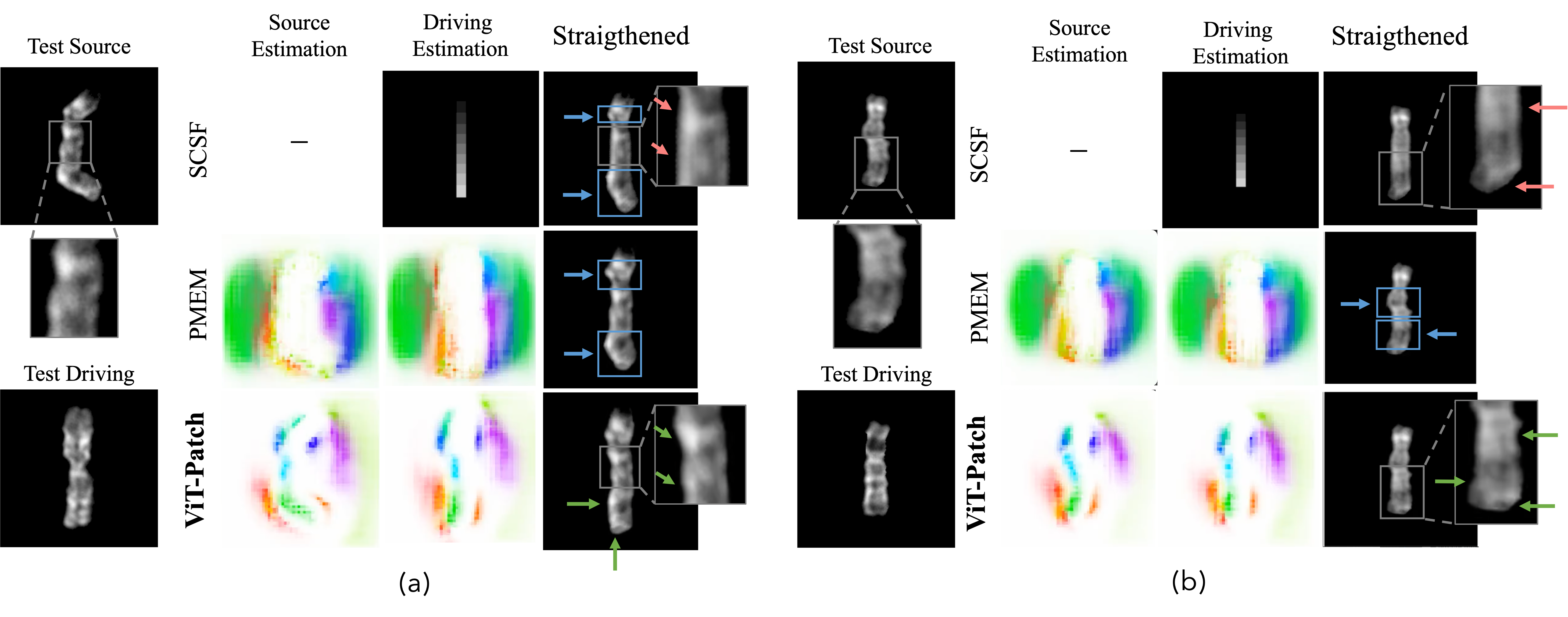}
	\vspace{-7pt}
	\caption{\footnotesize Visual results of chromosome straightening generated by different methods.} \label{Fig_sota}
\vspace{-5pt}
\end{figure*}

To alleviate this drawback, we propose a ViT-Patch discriminator (Fig. \ref{Fig_vitgan}-b) that encodes not only local details but also global feature connection. The final loss is the standard adversarial loss of the generator and discriminator in the cGAN-based framework~\cite{isola2017image}. During testing, the converged model straighten chromosomes by feeding \textit{test source} and \textit{test driving} images. It is worth noting that the \textit{test driving} can be a different straight chromosome to the \textit{test source} since the motion representation has been learned. 

\textbf{ViT-Patch Discriminator.} Fig.~\ref{Fig_vitgan}-b and Fig.~\ref{Fig_vitgan}-c give the comparison between our proposed ViT-Patch discriminator and a basic patch discriminator. In a basic patch GAN~\cite{isola2017image}, the semantic content of the corresponding patches at the same position between the source and generated images is generally the same. However, in this task, curvatures of a chromosome change after straightening. As a result, patches at the same position before and after straightening may contain significantly different chromosome patterns (\eg~the concatenated patches with black arrows in Fig.~\ref{Fig_vitgan}-b and Fig.~\ref{Fig_vitgan}-c). 

In training a basic patch GAN, two patches at the same position are concatenated and fed into the discriminator network ($D_P$) that contains several consecutive convolutional blocks to output the \textit{Feature Map2} for adversarial training (Fig. \ref{Fig_vitgan}-c). Although adjacent patches overlap according to the receptive field, long-distance patches are independent and not informatively linked. This deficiency may lead to inaccuracy and limited performance in the chromosome straightening task. To address this, for the proposed ViT-Patch discriminator, patches at the same position are fed into a convolutional patch embedding layer (\textit{C$_{PE}$}), and then processed by $N$ Multi-Head Self-Attention (MHSA) Blocks as ViT~\cite{dosovitskiy2020image} (4-16 used for ablation experiments in Section \ref{sec:abl}). Afterwards, the encoded features contain long-range dependencies across the image to compensate for information loss. Thus the \textit{Feature Map1} contains not only the local semantic content but also the information connectivity of the entire chromosome. 

\vspace{-7pt}
\subsection{{SL-matching} Scheme}
\label{sec:sl}

In addition to proposing a ViT-Patch discriminator to improve generalization, empirically, we found a large shape difference between the \textit{test source} and \textit{test driving} chromosome, leading to significant distortion. To alleviate this problem, we propose a Size-LPIPS matching scheme ({SL-matching}) to select a \textit{test driving} with a similar size and shape for each \textit{test source} image. The {SL-matching} scheme is comprised of two phases. In {Phase 1} we perform a line-by-line scan of the chromosome image and record the midpoint of each line containing non-zero pixels. The resulting set of midpoints is first smoothed using a moving average algorithm, and its {length} calculated. Next, we perform a column-by-column scan, where we take the \textit{x}-axis coordinates of all non-zero pixel columns as the {width}. The top three candidates whose {lengths} and {widths} are most similar to the corresponding \textit{test source} are selected. In {Phase 2} we calculate the perceptual score between a \textit{test source} and each candidate chromosome. This score is the average result generated by LPIPS with AlexNet and VGG backbones~\cite{zhang2018unreasonable}. Finally, the selected image has the largest size and perceptual similarity.

\section{Experiments}

\subsection{Experiment Setup}


\textbf{Datasets and Implementation Details.} A total of 16696 chromosome micrographs were used for the evaluation as provided by \cite{song2021novel}. All data had been desensitized, and patients' personal information has been removed. Since 642 out of the 16696 chromosomes were straightened by SCSF~\cite{song2021novel}, these 642 chromosomes were used as the dataset (\texttt{D$_{train}$}) for training and testing the ViT-Patch GAN in a 4:1 ratio. In the training stage, the original chromosomes and corresponding straightened results in \texttt{D$_{train}$} were utilized as \textit{training source} and \textit{training driving} images, respectively. Our clinical experts select 1200 real-world straight chromosomes of different lengths from all chromosome types as the dataset, \texttt{D$_{driving}$}. In the testing stage, \texttt{D$_{driving}$} was used to select \textit{test driving} for each \textit{test source} utilizing the {SL-matching} scheme. A large dataset was employed in this study to assess generalization capacity: the remaining 14854 chromosome images provided by~\cite{song2021novel} (\texttt{D$_{large}$}).

All experiments were implemented using one NVIDIA GeForce RTX 2080Ti GPU and coded using PyTorch. The Adam optimizer and a batch size of 1 were used. The total training epoch was 50 and the ratio of the training and test sets was 4:1 (5-fold cross-validation). The initial learning rates of the generator and discriminator were $5e^{-5}$ and $1e^{-5}$, respectively. We used MultiStepLR for the generator and discriminator with milestones 30 and 45. After completing cross validation, all 642 chromosomes of \texttt{D$_{train}$} were straightened. Subsequently, we performed downstream classification experiments (7 chromosome types) using initial learning rate of $4e^{-5}$ and ReduceLRonPlateau with patience 5 and early stopping protocol with patience 20. All input images were preprocessed to png format (256$\times$256).


\begin{table}\scriptsize
	\caption{\footnotesize Comparison with existing studies, the best and second best results highlighted in bold and underlines. The DCA result is the average score of each experiment with cross-validation.}\label{table_quality}
	\centering
	\resizebox{1\columnwidth}{!}{
	\begin{tabular}{l c c c c c c c }
		\hline
		DCA (\%) & FID$\downarrow$ & LPIPS$_{A}$$\downarrow$ & LPIPS$_{V}$$\downarrow$ & DCA$_{R34}$ & DCA$_{R50}$ & DCA$_{D169}$ \\
		\hline
		Original &-&-&-& 91.72 & 85.63 & 83.59  \\
		SCSF~\cite{song2021novel} & 54.90 & 0.1272 & 0.0974 & 90.63 & \underline{86.87} &  85.31 \\
		FOMM~\cite{siarohin2019first} & 53.71 & 0.1310 & 0.0982 & \underline{92.19} & 85.16 & \underline{88.28}  \\
		PMEM~\cite{siarohin2021motion} & \underline{43.80} & \underline{0.1196} & \underline{0.0897} & 92.03 & 86.09 & 84.69   \\
		\textbf{ViT-Patch} & \textbf{42.21} & \textbf{0.1160} & \textbf{0.0874} & \textbf{93.91} & \textbf{90.16} & \textbf{89.06}   \\
		\hline
	\end{tabular}
	}
	\scriptsize{\\$\downarrow$ represents that lower results are better.}\\
	\vspace{-8pt}
\end{table}

\textbf{Evaluation Metrics.} To quantitatively assess the quality of straightened chromosomes, we used the following three evaluation metrics: (i) {Fréchet Inception Distance} (FID) (ii) {Learned Perceptual Image Patch Similarity} (LPIPS) and (iii) {Downstream Classification Accuracy} (DCA). FID~\cite{heusel2017gans} assesses the quality of generated images by comparing the distribution between the real and generated images. LPIPS~\cite{zhang2018unreasonable} estimates the perceptual similarity. Its results are closer to human judgment than many traditional metrics, such as $\mathcal{L}_2$, PSNR and SSIM, especially for deformed objects (between original and straightened chromosomes)~\cite{zhang2018unreasonable}. LPIPS$_{A}$ and LPIPS$_{V}$ represent the scores generated by AlexNet and VGG backbones. Since chromosome classification is an important downstream task of chromosome straightening for subsequent karyotype construction and pathological diagnosis, the DCA can be used to assess the performance.

\vspace{-4pt}
\subsection{Comparison with State-of-the-Art Methods}

Table \ref{table_quality} gives the quantitative assessment of chromosome straightening on \texttt{D$_{train}$}. We compare the ViT-Patch GAN with state-of-the-art studies, SCSF~\cite{song2021novel}, FOMM~\cite{siarohin2019first} and PMEM~\cite{siarohin2021motion}. These results are the average scores using 5-fold cross-validation on \texttt{D$_{train}$}. The FID value of ViT-Patch GAN decreases from 54.90 to 42.21, implying that the straightening quality of our method is closest to the real-world chromosome images. Moreover, ViT-Patch GAN achieves the best LPIPS scores on both AlexNet and VGG backbones, demonstrating the clear advantages of our proposed method in this task.

\begin{figure}
	\centering
	\includegraphics[width=0.97\linewidth]{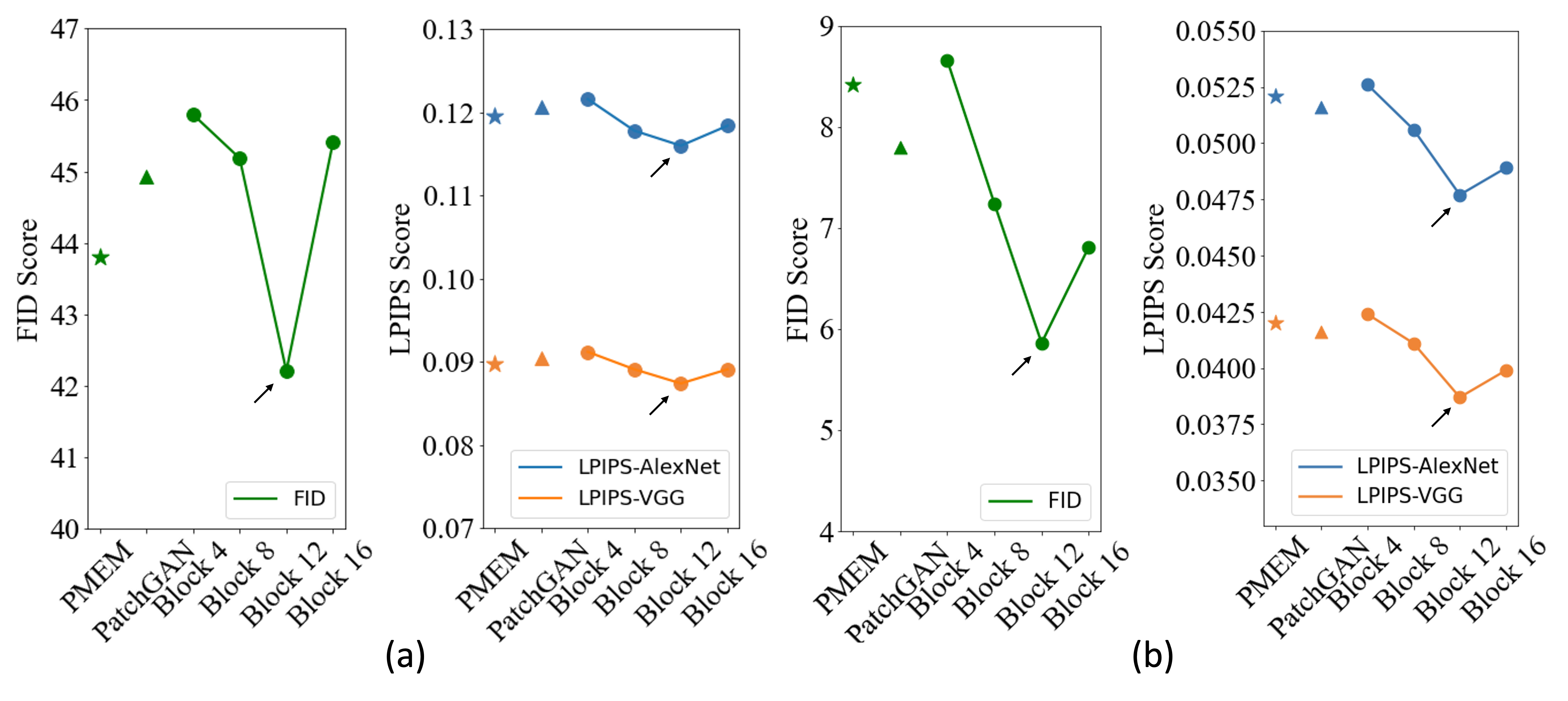}
	\caption{\footnotesize Comparison of straightening performance (a) of ablation study and (b) on a large-scale dataset (\texttt{D$_{large}$}).} \label{Fig_comb}
\vspace{-7pt}
\end{figure}

\begin{figure}
	\centering
	\includegraphics[width=0.88\linewidth]{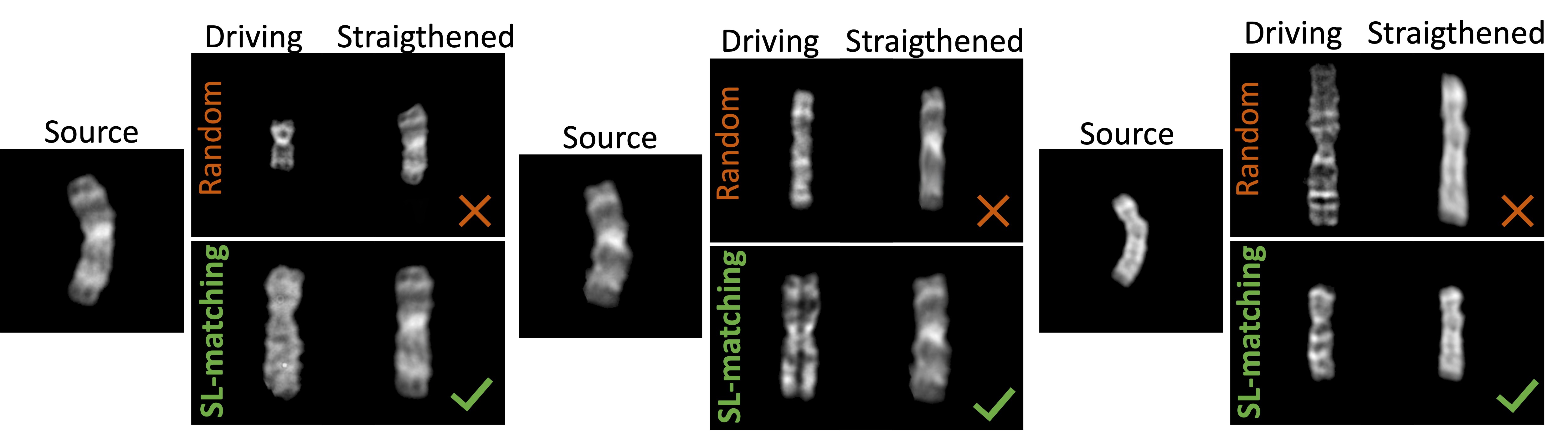}
	\caption{\footnotesize Comparison of straightening performance based on different \textit{test driving} images selected by random and our proposed {SL-matching} schemes.} \label{Fig_random_str}
	\vspace{-7pt}
\end{figure}

We perform downstream classification experiments using 642 chromosomes straightened by these state-of-the-art methods in exactly the same configuration. Each DCA is the average best result generated by 4-fold cross-validation classification experiments with commonly used models, ResNet34 (R34), ResNet50 (R50) and DenseNet169 (D169)~\cite{he2016deep,huang2017densely}. Although there is overfitting with the increasing depth of classification networks, we observe that ViT-Patch GAN significantly outperforms other methods by a large margin on DCA$_{R34}$, DCA$_{R50}$ and DCA$_{D169}$. The proposed method achieves gains of 2.19\%, 4.53\% and 5.47\% on DCA compared to the original bent chromosomes (baselines). Such results highlight that our proposed ViT-Patch GAN has an excellent reconstruction quality of straightened chromosomes.

Fig.~\ref{Fig_sota} gives the comparison between two straightened chromosomes. Compared to SCSF and PMEM, ViT-Patch GAN achieves the most accurate straightening details (green arrows) with fewer reconstruction errors. These errors are generally of two types: (i) bending is not well recovered (blue arrows with boxes), and (ii) the chromosome shape and details are not well preserved (red arrows). This inadequate straightening is mainly caused by inaccurate motion representation. For SCSF, only a backbone responsible for straightening results in a lack of features. The estimated regions of PMEM are wide and mostly located on the background, resulting in the curvature being almost not straightened (blue arrows). In contrast, ViT-Patch GAN estimates more meaningful and accurate motion representation for straightening, resulting in highly reliable straightening results. 


\vspace{-7pt}
\subsection{Ablation Study and Cross-Dataset Experiments}
\label{sec:abl}

We conduct ablation experiments to compare the performance of different architectures. We also implement PatchGAN, which uses a basic patch discriminator with PMEM as the generator. Blocks 4-16 are experiments for our proposed ViT-Patch discriminator with a series number of {MHSA} blocks. Fig. \ref{Fig_comb}-a shows that the performance on all metrics increases from Blocks 4 to 12, and Blocks 12 of our method achieves the best results (black arrows). In contrast, Blocks 16 may lead to overfitting, resulting in decreased performance. 

We further conduct experiments with the exactly same ViT-Patch GAN model (Blocks 12), but only with different \textit{test driving} images picked by random selection and the proposed {SL-matching} scheme. Fig.~\ref{Fig_random_str} demonstrates that the SL-matching scheme is important to drive source images for generating accurate straightening results. 

We perform cross-dataset experiments to evaluate the generalization capability of ViT-Patch GAN (Fig.~\ref{Fig_comb}-b). The models (those used in Section \ref{sec:abl}) are trained on \texttt{D$_{train}$}. The performance relationship between Fig.~\ref{Fig_comb}-b and Fig.~\ref{Fig_comb}-a is consistent. Our ViT-Patch GAN with Blocks 12 outperforms the others on all metrics, suggesting its robustness in the cross-dataset setting. Although our proposed chromosome straightening framework is trained on only 642 chromosomes (\texttt{D$_{train}$}), the ViT-Patch GAN learns the global motion representation rather than only mapping dependencies of specific patterns (\ie SCSF). The converged model can further straighten chromosomes of all types on the large dataset, \texttt{D$_{large}$}. Thus, it may be further utilized for large-scale applications of chromosome straightening. 

\vspace{-4pt}
\section{Conclusions}
\vspace{-3pt}
In this paper we propose a novel robust chromosome straightening framework, which includes a generator for learning chromosome motion representation and a ViT-Patch discriminator for generating more realistic straightened results. The ViT-Patch discriminator encodes both the local detail and long-range dependency. Qualitative and quantitative results demonstrate that the efficacy of our proposed method in retaining more chromosome shape and banding pattern details. Our framework also has excellent generalization capability in chromosome straightening for a large size dataset.

\small
\section{COMPLIANCE WITH ETHICAL STANDARDS}
\vspace{-7pt}
\label{sec:con}
\small
All procedures performed in studies involving human participants were in accordance with the ethical standards of the institutional and/or national research committee.

\vspace{-4pt}
\small
\section{ACKNOWLEDGMENTS}
\label{sec:ack}
\small
\vspace{-7pt}
This work was supported by the Key Program Special Fund in XJTLU (KSF-A-22).
\vspace{-9pt}
\bibliographystyle{IEEEbib}
\small
\bibliography{Manuscript}

\end{document}